\documentclass[runningheads]{llncs}
\usepackage[numbers,sort&compress]{natbib}
\usepackage{graphicx}
\usepackage{float} 
\usepackage{subfigure}
\usepackage{multirow}
\usepackage{amsmath}

\begin{document}
\title{Video-based Hierarchical Species Classification for Longline Fishing Monitoring}
\titlerunning{Video-based Hierarchical Species Classification}
\author{Jie Mei\inst{1}\orcidID{0000-0002-8421-7981} \and
Jenq-Neng Hwang\inst{1}\orcidID{0000-0002-8877-2421} \and
Suzanne Romain\inst{2}\and 
Craig Rose\inst{2}\and 
Braden Moore\inst{2}\and 
Kelsey Magrane\inst{2}}
\authorrunning{J. Mei et al.}
\institute{University of Washington, Seattle, WA 98195, USA\\
\email{\{jiemei, hwang\}@uw.edu}\\
\url{https://ipl-uw.github.io/} \and
EM Research and Development, National Oceanic and Atmospheric Administration (NOAA) Affiliate, Pacific States Marine Fisheries Commission, Seattle, WA 98115, USA\\
\email{\{suzanne.romain, craig.rose, braden.j.moore, kelsey.magrane\}@noaa.gov}}
\maketitle              % typeset the header of the contribution
\begin{abstract}The goal of electronic monitoring (EM) of longline fishing is to monitor the fish catching activities on fishing vessels, either for the regulatory compliance or catch counting. Hierarchical classification based on videos allows for inexpensive and efficient fish species identification of catches from longline fishing, where fishes are under severe deformation and self-occlusion during the catching process. More importantly, the flexibility of hierarchical classification mitigates the laborious efforts of human reviews by providing confidence scores in different hierarchical levels. Some related works either use cascaded models for hierarchical classification or make predictions per image or predict one overlapping hierarchical data structure of the dataset in advance. However, with a known non-overlapping hierarchical data structure provided by fisheries scientists, our method enforces the hierarchical data structure and introduces an efficient training and inference strategy for video-based fisheries data. Our experiments show that the proposed method outperforms the classic flat classification system significantly and our ablation study justifies our contributions in CNN model design, training strategy, and the video-based inference schemes for the hierarchical fish species classification task.

\end{abstract}
\section{Introduction}
\subsection{Electronic Monitoring (EM) of Fisheries}
Automated imagery analysis techniques have drawn increasing attention in fisheries science and industry \cite{zion2012use, guptatrends, white2006automated, chuang2013aggregated,huang2016chute,williams2016automated, chuang2014tracking,huang2016live,wang2016closed}, because they are more scalable and deployable than conventional manual survey and monitoring approaches.

One of the emerging fisheries monitoring methods is electronic monitoring (EM), which can effectively take advantage of the automated imagery analysis for fisheries activities \cite{huang2016live}. The goal of EM is to monitor fish captures on fishing vessels either for catching counting or regulatory compliance. Fisheries managers need to assess the amount of fish caught by species and size to monitor catch quotas by vessel or fishery. Such data are also used in analyses to evaluate the status of fish stocks. Managers also need to detect the retention of specific fish species or sizes of particular species that are not allowed to be kept. Therefore, accurate detection, segmentation, length measurement, and species identification are critically needed in the EM systems. 

\subsection{Hierarchical Classification}
Especially in the EM systems, a hierarchical classifier is more meaningful for the fisheries than a flat classifier with the standard softmax output layer. The hierarchical classifier can predict coarse-level groups and fine-level species at the same time. If the system predicts some images with high confidence in one coarse-level group but with low confidence in the corresponding fine-level species, then a hierarchical classifier stops predictions of those images at the correct coarse-level group and allows fisheries personnel to assign corresponding experts to review those images and get the correct fine-level labels.

\begin{figure}
\centering\includegraphics[width=0.6\textwidth]{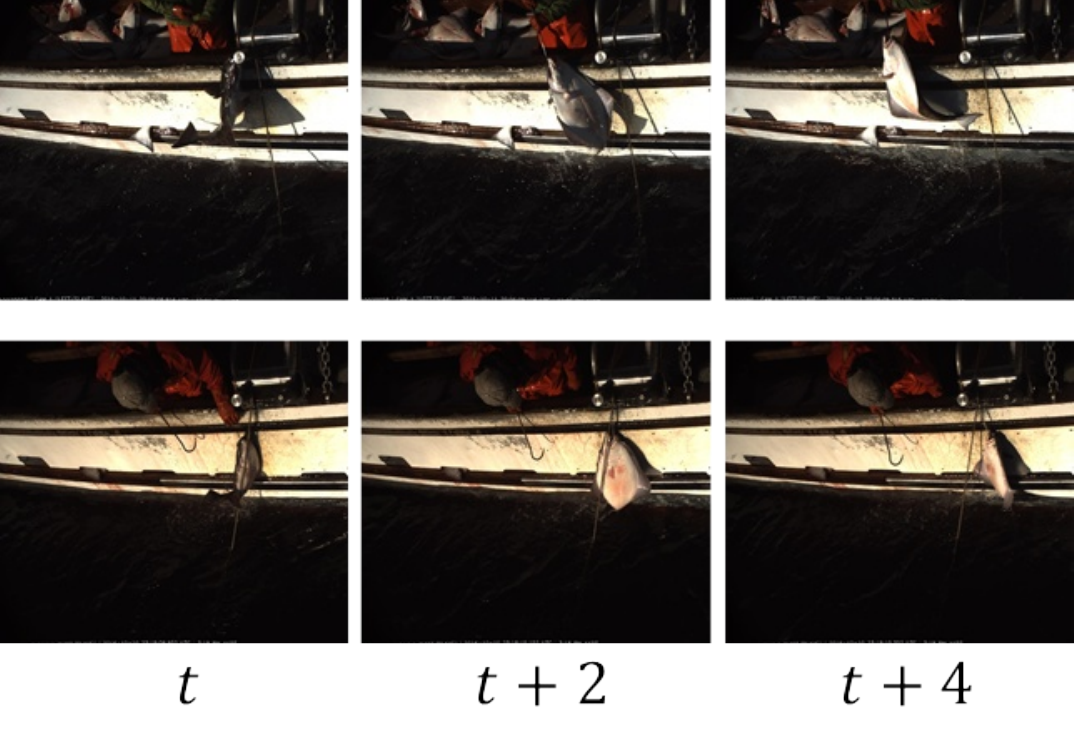}
\caption{Longline Fishing: Each column is a sequence of an individual fish caught on a longline hook, as it is being pulled up from the sea and over the rail of the fishing vessel.} \label{vessel}
\end{figure}

To address the hierarchical classification needs, in this paper, we develop a video-based hierarchical species classification system for the longline fishing monitoring, where fish are caught are caught on hooks and viewed as they are pulled up from the sea and over the rail of the fishing vessel as shown in Fig.\ref{vessel}.

The proposed hierarchical prediction, which allows coarse-level prediction to be the final output if fine-level confidence score is too low, improves accuracy on tail-class species when training data follows a long-tail (imbalanced) distribution. Our contributions can be summarized as follows: 1) Our proposed CNN architecture enforces an effective hierarchical data structure. 2) An efficient training strategy. 3) Two robust video-based hierarchical inference schemes.

The remaining sections of this paper are organized as follows. In Section II, overviews of the related works for flat classifiers with the standard softmax output layer and hierarchical classifiers are provided. Section III describes the proposed system in details. The experimental results, including the ablation study, are demonstrated and discussed in Section IV. Finally, Section V gives the conclusion of this work.

\section{Related Work}
\subsection{Flat Classifiers}
We use 'flat classifiers' to represent all deep learning classification systems with softmax as the final layer to normalize the outputs of all classes, without introducing any hierarchical level of prediction. 

 AlexNet \cite{krizhevsky2012imagenet} is the first CNN-based winner in 2012 ImageNet Large Scale Visual Recognition Challenge (ILSVRC), which introduces the 1000-way softmax layer for classifying the 1000 classes of objects. The subsequent ILSVRC winners, VGGNet \cite{simonyan2014very}, GoogLeNet \cite{szegedy2015going}, and ResNet \cite{he2016deep} continue to use softmax as the final layer to achieve good performance. Until now, flat classifiers with softmax operations as the final layer are the dominant design structure for classification tasks.

\subsection{Hierarchical Classifier}
A hierarchical classifier means the system can output all confidence scores at different levels in the hierarchical data structure. One obvious advantage is that if the confidence score of a sample is too low at the fine level but very high at coarse level, then we can use the coarse-level prediction to be the final prediction. In contrast, flat classifiers have no alternative ways if the confidence score is too low at the final prediction.

Hand-crafted features are used in \cite{huang2015hierarchical} for hierarchical fish species classification. Hierarchical medical image classification \cite{kowsari2020hmic} and text classification \cite{kowsari2017hdltex} use cascaded flat classifiers to be their hierarchical classifiers, which use only one flat classifier for each level's prediction. They stack CNN-based models with flat classifiers without considering any hierarchical architecture design and increased computational complexity. HDCNN \cite{Yan_2015_ICCV} introduces confidence-score multiplication operations to enforce hierarchical data structure but the model uses the same feature maps for both coarse level and fine level, resulting in learning an overlapping hierarchy of training data. B-CNN \cite{zhu2017b} uses different feature maps for different levels' predictions without enforcing any hierarchical data structure in the architecture. Deep RTC \cite{wu2020solving} adopts hierarchical classification to deal with long-tailed recognition, resulting in improved accuracy of tail classes. It adopts a simple confidence-score thresholding method, which is also adopted in our approach, to decide to output fine-level prediction or coarse-level prediction. But Deep RTC predicts an overapping hierarchical data structure in the first place, which is different with our situation.

\section{Proposed Method}

\subsection{Hierarchical Dataset} 
The hierarchical dataset utilized for training our system is professionally labeled and provided by the Fisheries Monitoring and Analysis (FMA) Division, Alaska Fisheries Science Center (AFSC) of NOAA, researchers can contact AFSC directly about getting permission to access this dataset and the corresponding hierarchical data structure.

\begin{figure} 
\centering
\includegraphics[width=1\textwidth]{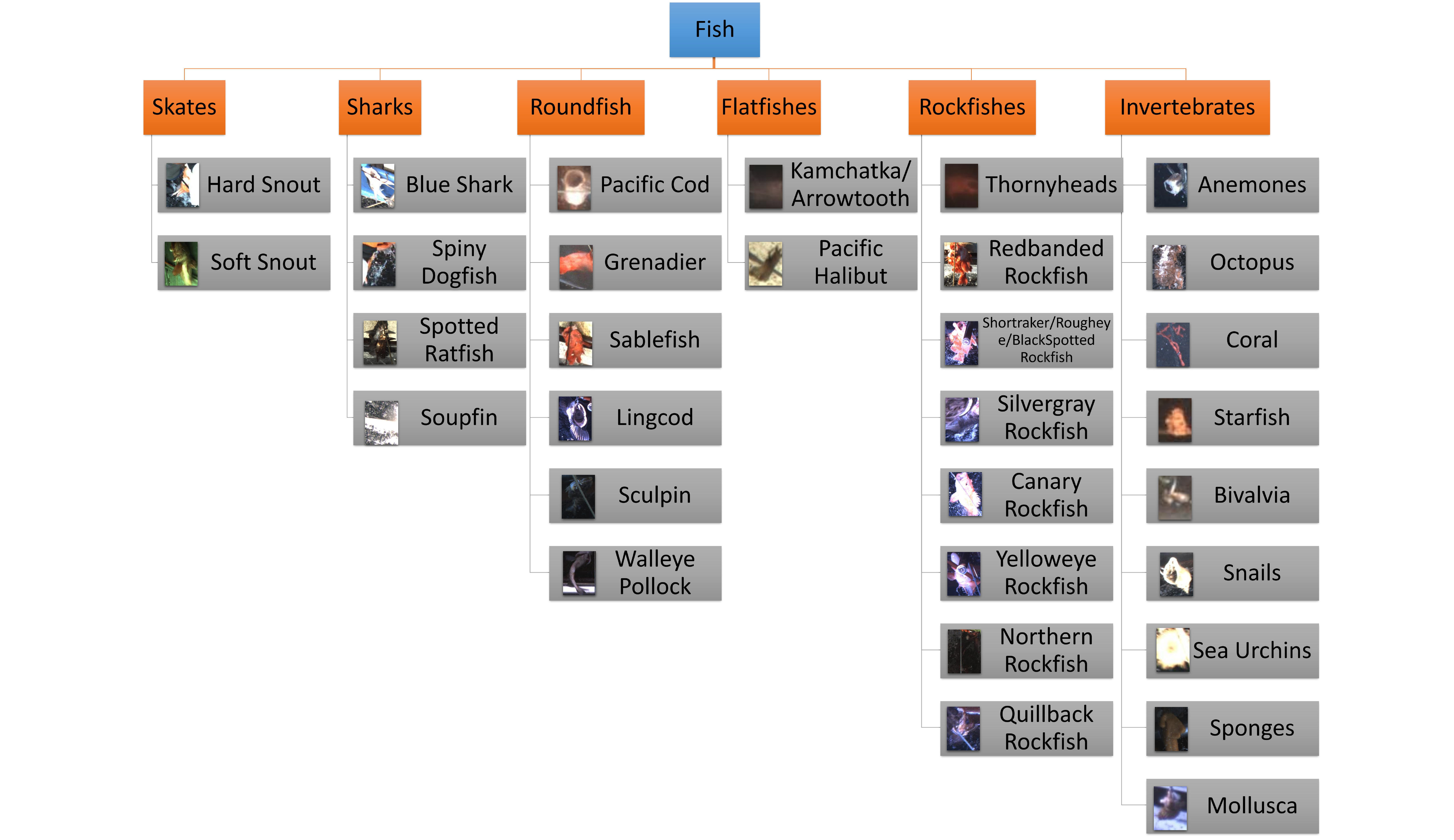}
\caption{Hierarchical Data Structure: The dataset, labeled and provided by NOAA fisheries scientists, includes frames and corresponding labels which are bounding box location, start and end frames' IDs of each individual fish, coarse-level group ground truth, and fine-level species ground truth. The sample images shown here are randomly chosen from the dataset.} \label{hierarchical structure}
\end{figure}

\begin{figure}
\centering  %图片全局居中
\subfigure[Image-Species Distribution]{
\label{Data img}
\includegraphics[width=0.48\linewidth]{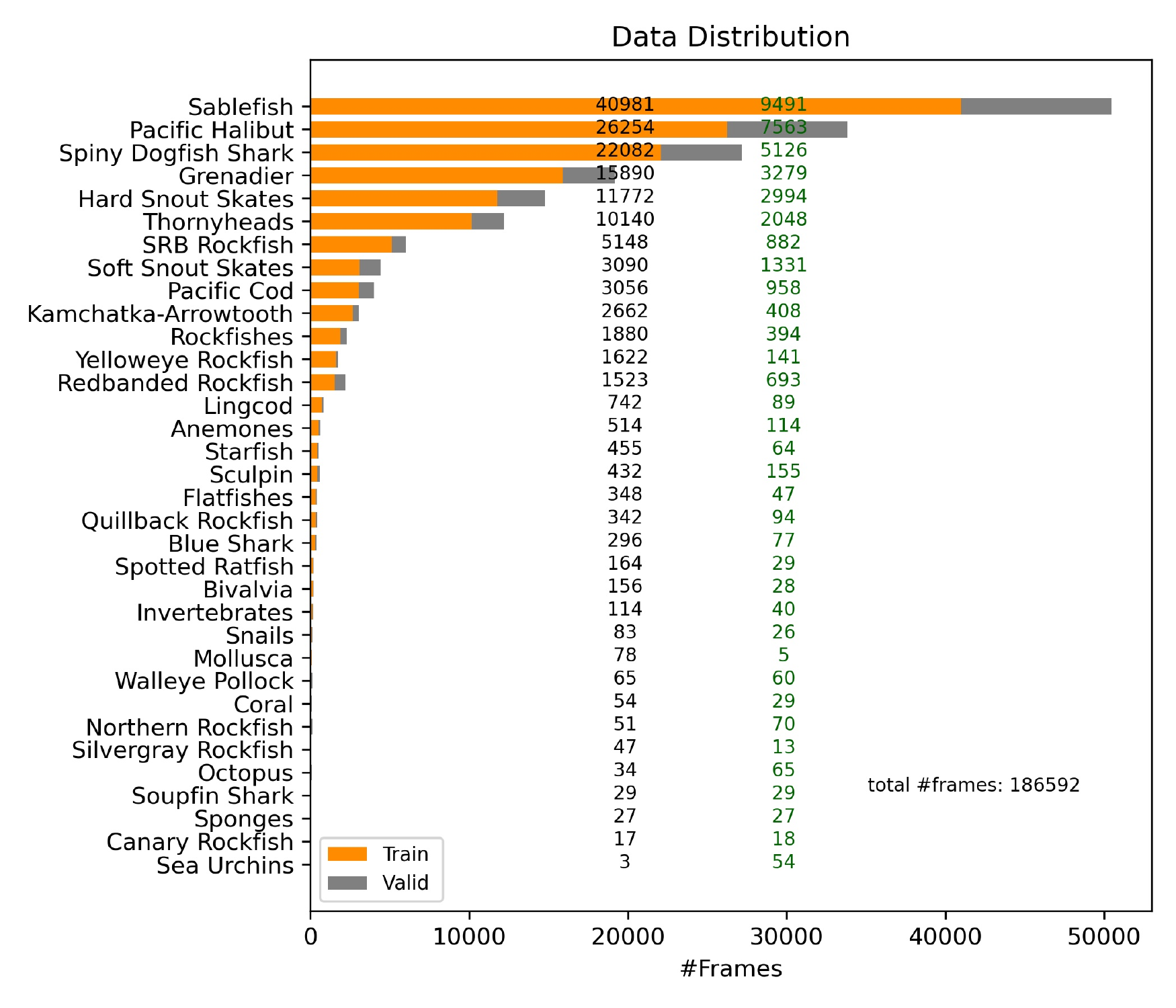}}
\subfigure[Track-Species Distribution]{
\label{Data Track}
\includegraphics[width=0.48\linewidth]{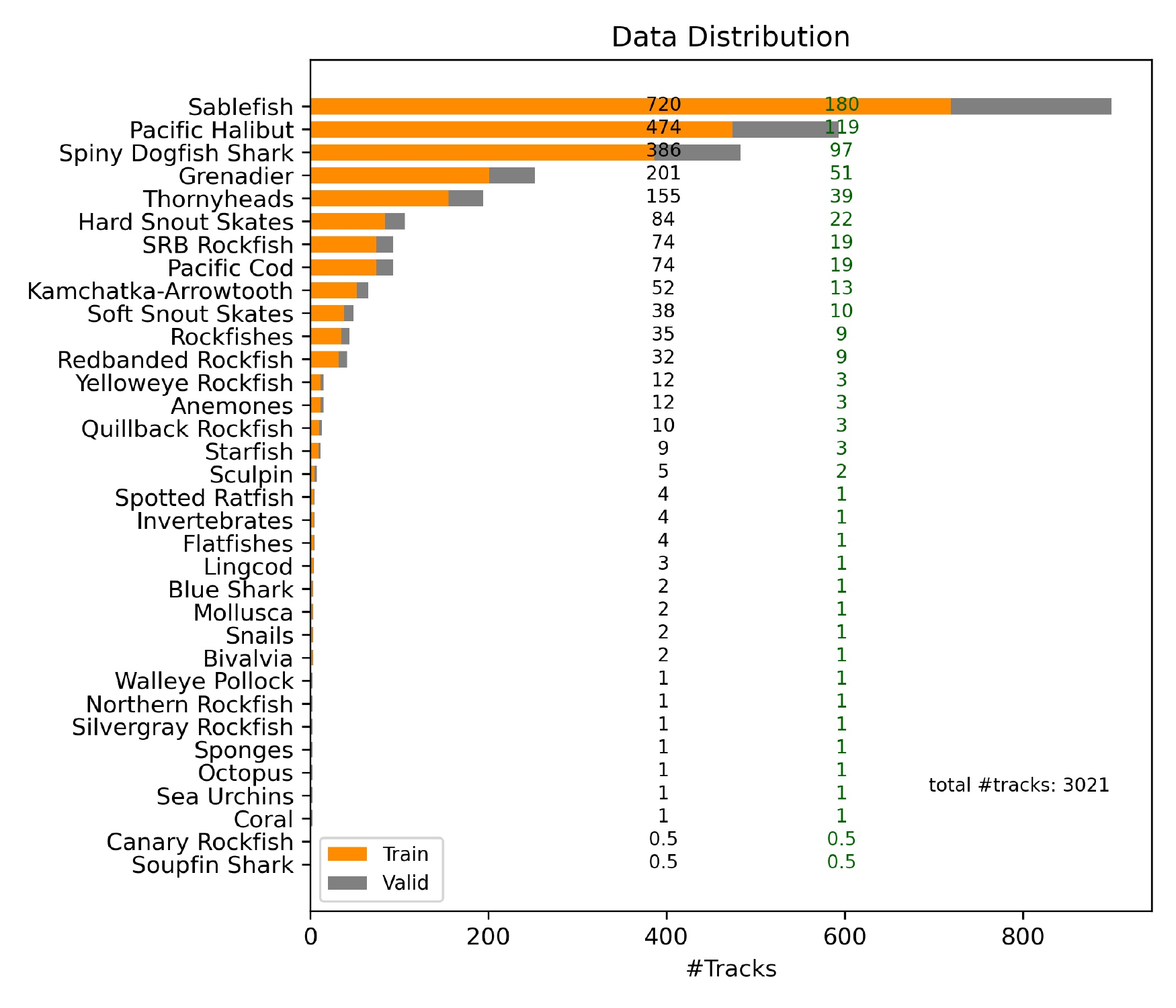}}
\caption{Dataset Distribution: The left column black numbers in both figures are the number of images or tracks for training while the green numbers are for evaluation. There is a $0.5$ in (b) because of only one track in the whole dataset, therefore we have to split it into $2$ tracks and denote each as $0.5$ track. 'SRB Rockfish' represents Shortraker/Rougheye/BlackSpotted Rockfish. Our dataset follows a long-tail (imbalanced) distribution.} 
\label{Data}
\end{figure}

To construct the dataset used for our system, we use labels of bounding box location to crop objects from raw videos and use labeled start and end frames' IDs for each individual fish to divide raw videos into individual tracks (video clips). There are 6 coarse-level groups and 31 fine-level species in this hierarchical dataset (see Fig.~\ref{hierarchical structure}). Our dataset is challenging because some fine-level species are very similar to one another. The total number of frames is 186,592 (see Fig.~\ref{Data img}). The total number of video clips/tracks is 3,021 (see Fig.~\ref{Data Track}). Each video clip contains one individual fish pulled up from sea surface to the fishing vessel during the longline fishing activities.

\subsection{Hierarchical Architecture} 
\begin{figure}
\includegraphics[width=\textwidth]{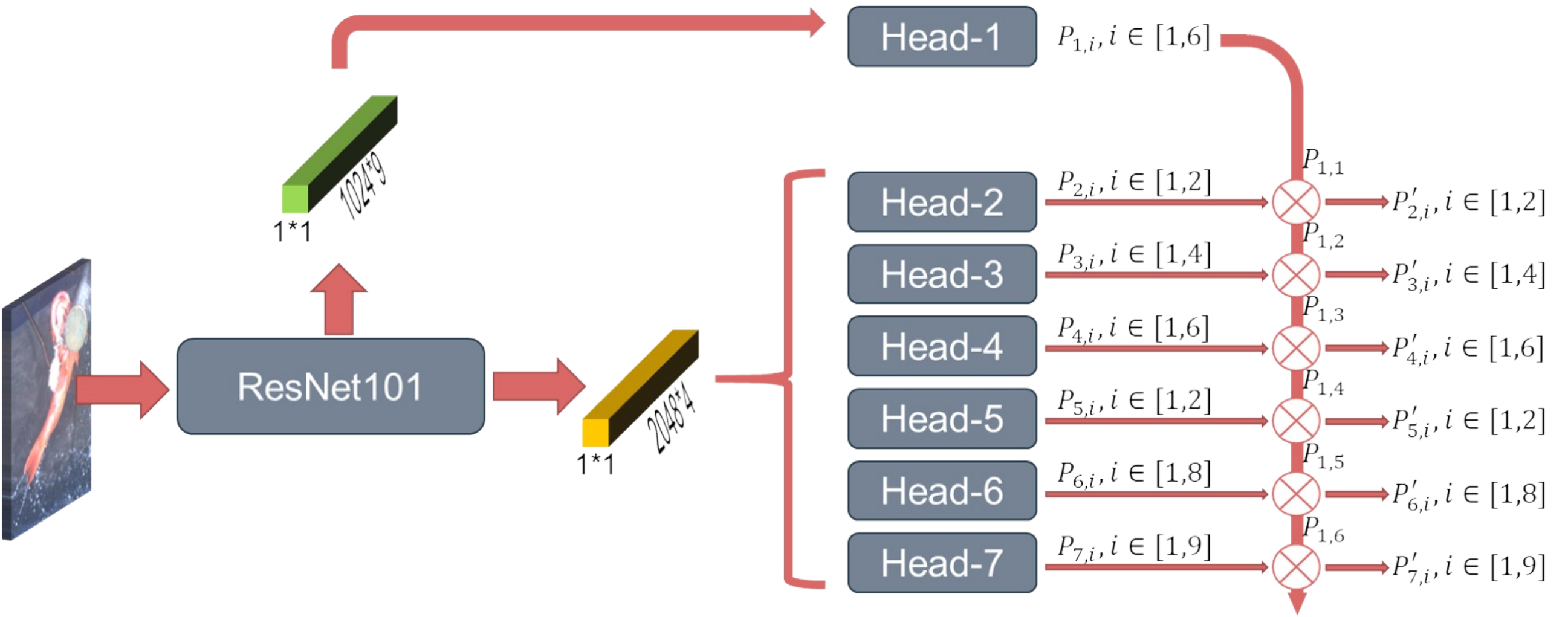}
\caption{Hierarchical Architecture: We call our 7 classification heads 'Hierarchical Heads'. Head-1 is for 6 groups in coarse-level and uses shallower feature maps extracted from the backbone for predictions, while the rest of 6 heads are for fine levels, i.e., Head-2 for 'Skates' group, Head-3 for 'Sharks' group, Head-4 for 'Roundfish' group, Head-5 for 'Flatfishes' group, Head-6 for 'Rockfishes' group, and Head-7 for 'Invertebrates' group, respectively. All fine-level heads use the shared deeper feature maps from the same backbone for predictions. Head-1 has two fully connected layers followed by a sofmax layer. The rest 6 fine-level heads have one fully connected layer followed by a sofmax layer.} \label{pipeline}
\end{figure}

Instead of using cascaded flat classifiers in our species identification in the longline fishing, as inspired by the success of Mask-RCNN \cite{DBLP:journals/corr/HeGDG17}, which feeds shared feature maps extracted from the backbone to different heads for object classification and instance segmentation at the same time, our proposed architecture is also an end-to-end training network including two parts: a backbone and several hierarchical classification heads (see Fig.~\ref{pipeline}). Inspired by B-CNN \cite{zhu2017b}, we use ResNet101 as our backbone to extract shallow feature maps from images for 'Head-1' and shared deeper feature maps for the other 6 classification heads. We use Head-1 for coarse-level (6 groups) classification and Head-2 to Head-7 for fine-level (31 species in total) predictions.

\subsubsection{Enforcing Hierarchical Data Structure} We use confidence-score multiplication operations to enforce the hierarchical data structure in our system. The final confidence score of one specific species is the product of the confidence score in the corresponding coarse group and the confidence score in that specific (fine) species as shown in the following equation, i.e.,

\begin{equation}
P_{j,i}^{\prime}=P_{1, j-1} \cdot P_{j, i}, \;\; j \in [2,7],
\end{equation}
where $P_{1, j-1}$ represents the confidence score in the $(j-1)^{th}$ group in coarse level, $P_{j, i}$ represents the confidence score in the $i^{th}$ species of the $(j-1)^{th}$ group, $P_{j,i}^{\prime}$ represents the final confidence score in the $i^{th}$ species of the $(j-1)^{th}$ group. As a result, the final confidence score, $P_{j,i}^{\prime}$, includes both scores from coarse level and fine level so that this CNN architecture enforces a hierarchical data structure when using the final confidence score product to calculate the training loss. Training loss is meaningful when using $P_{j,i}^{\prime}$ because the final layers in all heads are softmax outputs, which satisfy the following equations:

\begin{equation}
% \begin{aligned}
\left\{\begin{array}{r}
\sum_{i} P_{j, i} = 1, \\
\sum_{j=2}^{7} \sum_{i} P_{j,i}^{\prime} = 1.
\end{array}\right.
% \end{aligned}
\end{equation}

\subsubsection{Efficient Training Strategy} During the image-based training, one input image has both labeled coarse-level ground truth as well as fine-level ground truth. For our architecture, there are two options experimented about how to use these two ground truths. 

The first option is training the 'Head-1' and the fine-level head corresponding to the ground truth coarse-level group. Since the corresponding fine-level head is picked by coarse-level group ground truth, therefore the losses are only calculated based on these two heads.

\begin{equation}
Loss_{1} = -\sum_{i} y_{1,i} \cdot log(P_{1,i}) -\sum_{i} y_{2,i} \cdot log(P_{j,i}^{\prime}),
\end{equation}
where the first summation is the cross entropy loss in the 'Head-1' and $y_{1,i}$ is coarse-level ground truth. The second summation is the cross entropy loss in the corresponding fine-level head using final predictions, $P_{j,i}^{\prime}$, after confidence-score multiplication operations. Note that $y_{2,i}$ is the ground-truth among species within this fine-level head. This regular loss does not involve $P_{j i}^{\prime}$ from other heads, therefore, it does not fully enforce hierarchical data structure during training and only trains two heads each time.

The second option is training the 'Head-1' and all the other fine-level heads using final predictions $P_{j i}^{\prime}$ after confidence-score multiplication operations, resulting in a more efficient training strategy because it enables confidence-score multiplication operations to fully enforce hierarchical data structure during training and all heads can be trained simultaneously. 
\begin{equation}
Loss_{2} = -\sum_{i} y_{1,i} \cdot log(P_{1,i}) -\sum_{j=2}^{7} \sum_{i} y_{2,i}^{\prime} \cdot log(P_{j,i}^{\prime}),
\end{equation}
where $y_{2,i}^{\prime}$ denotes the ground-truth among 31 species. The reason why given one input image we can calculate cross entropy on all final predictions, $P_{j,i}^{\prime}$, is that  after the confidence-score multiplication operations, summation of these products is still 1.

\subsubsection{Video-based Inference Schemes} Although we use image-based training, where training loss is calculated on each individual input image, two video-based (track-based) inference methods are implemented and compared. Since for each input image frame, our system outputs confidence scores of 31 species, $P_{j,i}^{\prime}$, therefore the first way is to pick the species with the maximum average confidence score of all frames in each track to be the prediction for each track. 

The second way is to pick the species with maximum confidence score for every frame in each track, then uses majority vote to select one species as the prediction for that track. Finally, we calculate the average confidence scores among frames corresponding to the selected species. We report performance under these two video-based inference schemes and calculate their average confidence scores with image-based confidence scores in the following section. These two inference schemes can be summarized into the following equations: 

\begin{equation}
\left\{\begin{array}{c}
p_{1,i} = 1/T \cdot \sum_{t} P_{1,i,t}^{\prime},\\
p_{2,i} = 1/T \cdot \sum_{t} P_{j,i,t}^{\prime}, \;\; j \in [2,7],
\end{array}\right.
\end{equation}
where $t$ is frame index. $P_{j,i,t}^{\prime}$ is $P_{j,i}^{\prime}$ at the $t^{th}$ frame. In the first way, $T$ is the total number of frames in one video clip (a track from the start-frame to the end-frame of one catching), while in the second way, it is the total number of frames corresponding to the selected species in one video clip. As a result, $p_{1,i}$ is video-based average confidence scores in 6 groups and $p_{2,i}$ is video-based average confidence scores in 31 species.

\section{Experiments \& Discussion}
\subsection{Data Split}
We use the video-based data split, i.e., each short video clip (a track) is associated with one individual fish and all frames from $80\%$ of all tracks are used as training data for image-based training. All frames from the rest $20\%$ tracks are the evaluation data (see Fig.~\ref{Data}). As a result, images for training and evaluation are totally from tracks of different individual fishes. All hyper-parameters like training epochs, learning rate, data augmentation, and so on are kept the same in the following different competing approaches.

\subsection{Baseline}
The dominant species classification architecture is extracting deep features using CNN followed by a flat classifier. As a result, for the baseline, we use ResNet101 as the backbone and two fully connected layers followed by a 31-way softmax layer as the flat classifier head, which is a classic deep learning classification architecture. During training, we only use fine-level ground truth to calculate the cross-entropy loss based on the flat classifier output confidence scores in 31 species without any coarse-level predictions. \textit{From Table~\ref{experiments}, we can see the accuracy of the baseline is far below our hierarchical method}.

\subsection{Evaluation Methods}
Using all frames from the rest $20\%$ tracks for evaluation, we try the following evaluation methods, where we calculate both image-based accuracy as well as video-based (track-based) accuracy, denoted in the 'Unit' column in Table~\ref{experiments}. 

We also calculate classification accuracy on the coarse level based on coarse-level ground truth, denoted as 'Level-1' in Table~\ref{experiments}. 

% Third, for each image or video clip, given coarse-level ground truth, we can pick corresponding fine-level prediction and calculate accuracy on that fine-level independently, denoted as 'Level-2 A' in the Table~\ref{experiments}. 

Moreover, with confidence scores in the coarse level as well as the fine level, we can pick the species with maximum fine-level confidence score under the group with maximum coarse-level confidence score as the final prediction, which is denoted as 'Level-2 A' in Table~\ref{experiments}. While, with final confidence scores in 31 species, we can directly pick the species with the maximum confidence score product of coarse and fine levels as the final prediction, which is denoted as 'Level-2 B' in Table~\ref{experiments}. For these two metrics ('Level-2 A' and 'Level-2 B') in video-based schemes, we further use either maximum average confidence score (denoted as '$video$') or majority vote (denoted as '$video^{*}$') to report the performance, as discussed in the 'Video-based Inference Schemes' in Section 3.2. 

Finally, with these final confidence scores, $P_{j,i}^{\prime},\;\;j \in [2,7]$, in 31 species, for image-based inference, we can also decide to go back to the coarse level if the maximum confidence score product is below a threshold and calculate the accuracy on the coarse level for that input, otherwise stay at the fine level and calculate the accuracy for that input. For video-based inference methods, we compare the average confidence score mentioned in 'Video-based Inference Scheme' section with the threshold. This metric, being able to stay at a coarse level, is denoted as 'Level-2 C' in Table~\ref{experiments}. Theoretically, the ceiling limit of 'Level-2 C' is 'Level-1' if all samples stop at the coarse level. Therefore we use the greedy search to find a threshold for each scheme in Table~\ref{experiments} to make sure that after stopping at the coarse level, the overall video-based inference accuracy will not degrade. We fix these thresholds in image-based inference for every competing scheme.

\begin{table}
\centering
\caption{Comparison with Flat Classifier and Ablation Study: '$video$' denotes video-based inference by using average confidence score among 31 species to pick one predicted species for each track. '$video^{*}$' denotes video-based inference through majority vote to pick one species for each track. Two numbers under 'Level-2 C' column following the accuracy value are total number of stopping at coarse-level and total number of proceeding to fine-level respectively.}
\label{experiments}
\begin{tabular}{c|c|c|c|c|l}
\hline
Model &  Unit & Level-1 & Level-2 A & Level-2 B & \multicolumn{1}{c}{Level-2 C}\\
\hline
Baseline &  img & - & - & 78.3 & \multicolumn{1}{c}{-}\\
\hline
\multirow{3}{*} {Scheme-1} & img & 86.3 & 77.4 & 77.4 & 82.0(8567, 27393)\\
& $video^{*}$ & 93.2 & 86.5 & 86.6 & 93.2(298, 319)\\
& $video$ & 93.4 & 86.5 & 86.8 & 93.4(293, 324)\\
\hline
\multirow{3}{*} {Scheme-2} & img & 88.4 & 79.9 & 80.0 & 84.6(8660, 27300)\\
& $video^{*}$ & 94.3 & 88.6 & 88.9 & 94.3(329, 288)\\
& $video$ & 94.9 & 88.9 & 88.8 & 94.9(328, 289)\\
\hline
\multirow{3}{*} {Scheme-3} & img & 91.0 & 82.3 & 82.3 & 86.3(5830, 30130)\\
& $video^{*}$ & 96.3 & 90.6 & 90.3 & 96.3(286, 331)\\
& $video$ & \textbf{96.4} & \textbf{90.9} & \textbf{90.9} & \textbf{96.4}(293, 324)\\
\hline
\end{tabular}
\end{table}

\textit{From Table~\ref{experiments}, we can see video-based inference is always better than image-based inference in all competing schemes}. And these two video-based inference methods, average confidence and majority vote, are comparable with each other.

Scheme-3 is our full system demonstrated in Fig.~\ref{pipeline}, which includes confidence-score multiplication operations to enforce hierarchical data structure and uses the efficient training strategy ($Loss_2$). Scheme-2 only removes the efficient training strategy and uses $Loss_1$ instead. Scheme-1 removes confidence-score multiplication operations in the architecture but keeps 7 heads. It also removes the efficient training strategy and instead uses standard cross-entropy losses on 'Head-1' and the fine-level head corresponding to the ground truth coarse-level group. Scheme-1 shares the same architecture as B-CNN \cite{zhu2017b}. When evaluating under 'Level-2 B' and 'Level-2 C' for Scheme-1, we have to multiply the coarse-level confidence scores with the fine-level confidence score in advance to get the final confidence scores. 

\begin{figure}[t]
\centering  %图片全局居中
\subfigure[Precision at Level-1]{
\label{level-1 acc}
\includegraphics[width=0.48\linewidth]{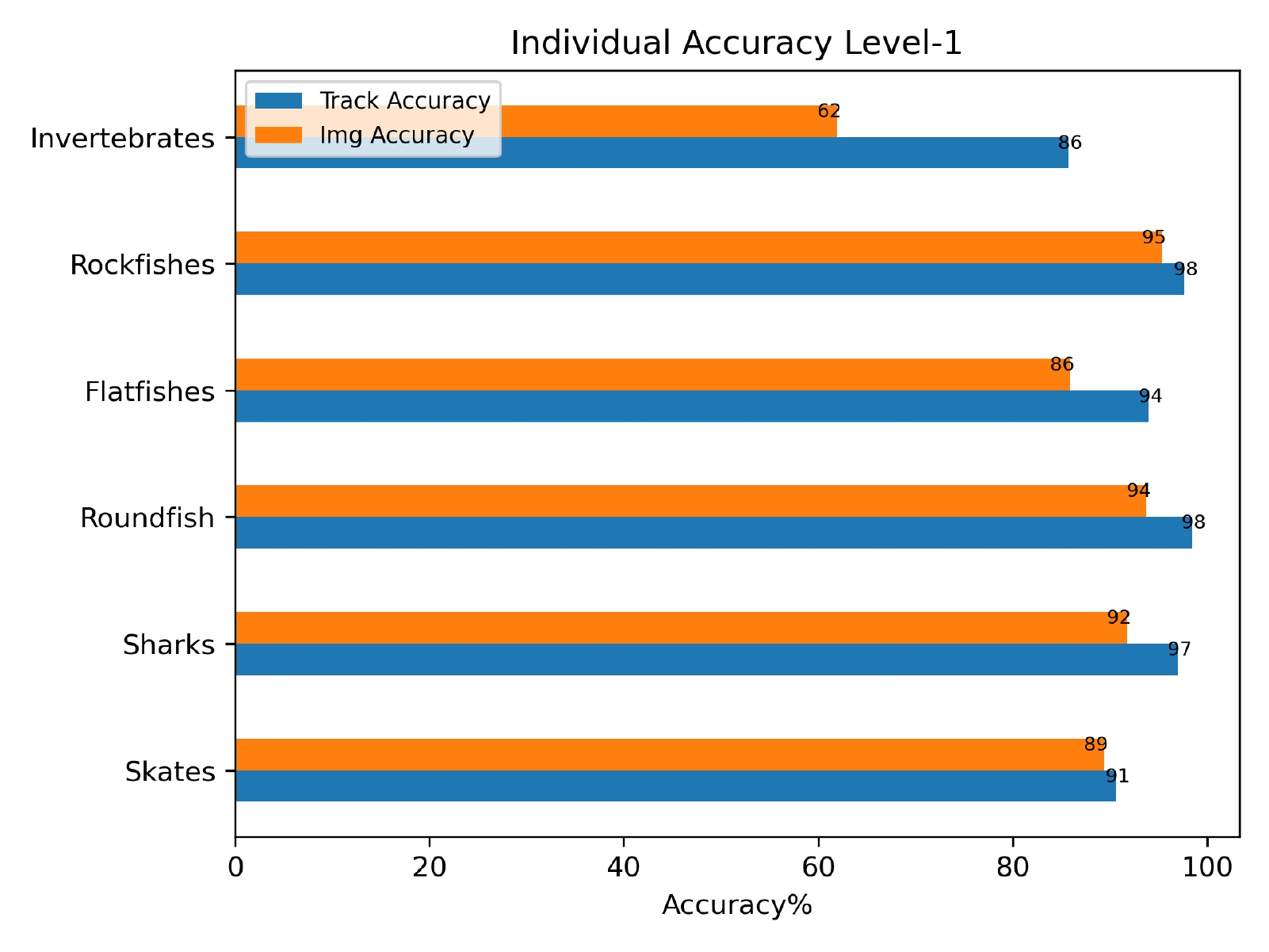}}
\subfigure[Precision at Level-2 A]{
\label{level-2 A}
\includegraphics[width=0.48\linewidth]{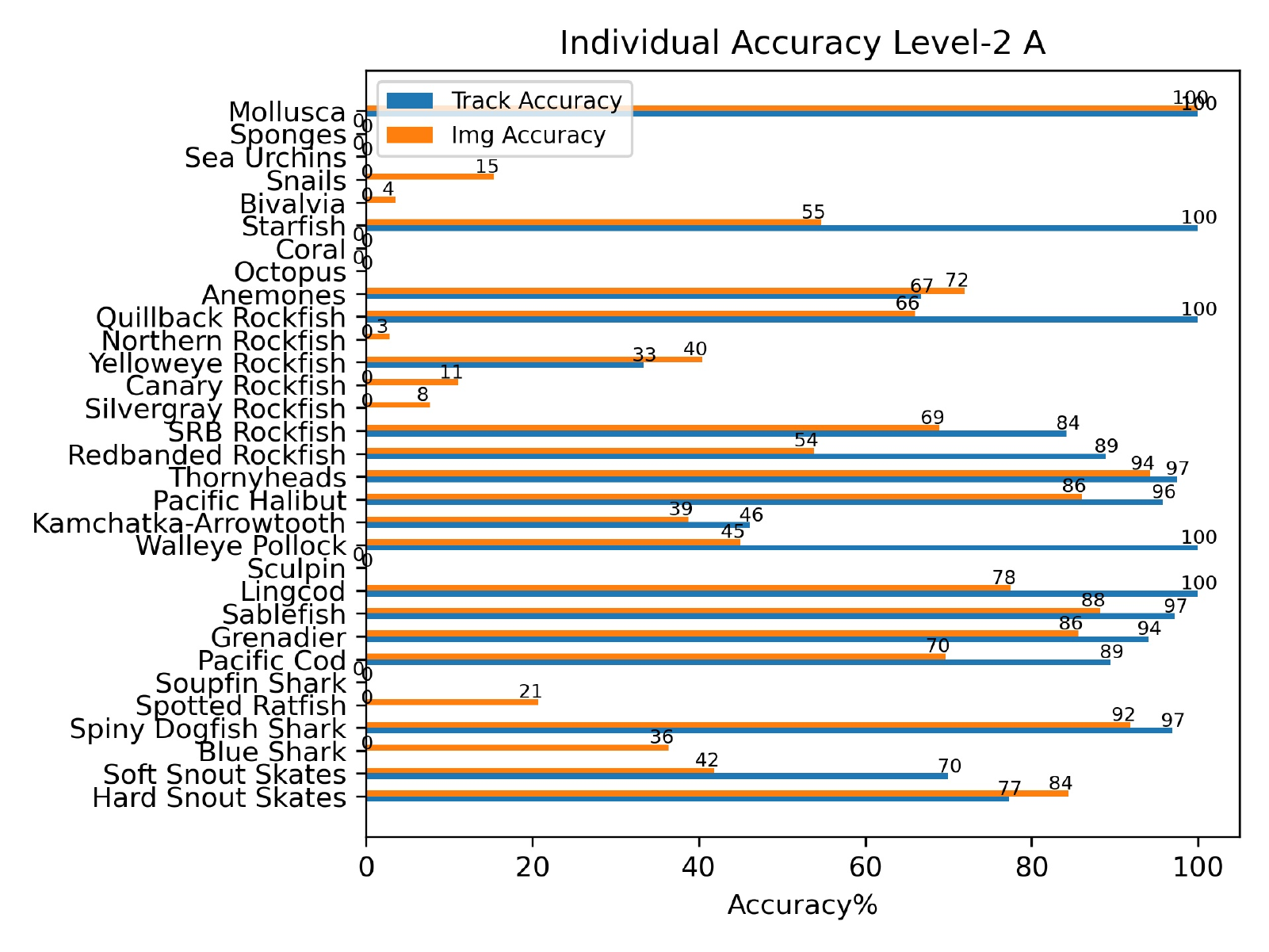}}
\subfigure[Precision at Level-2 B]{
\label{level-2 B}
\includegraphics[width=0.48\linewidth]{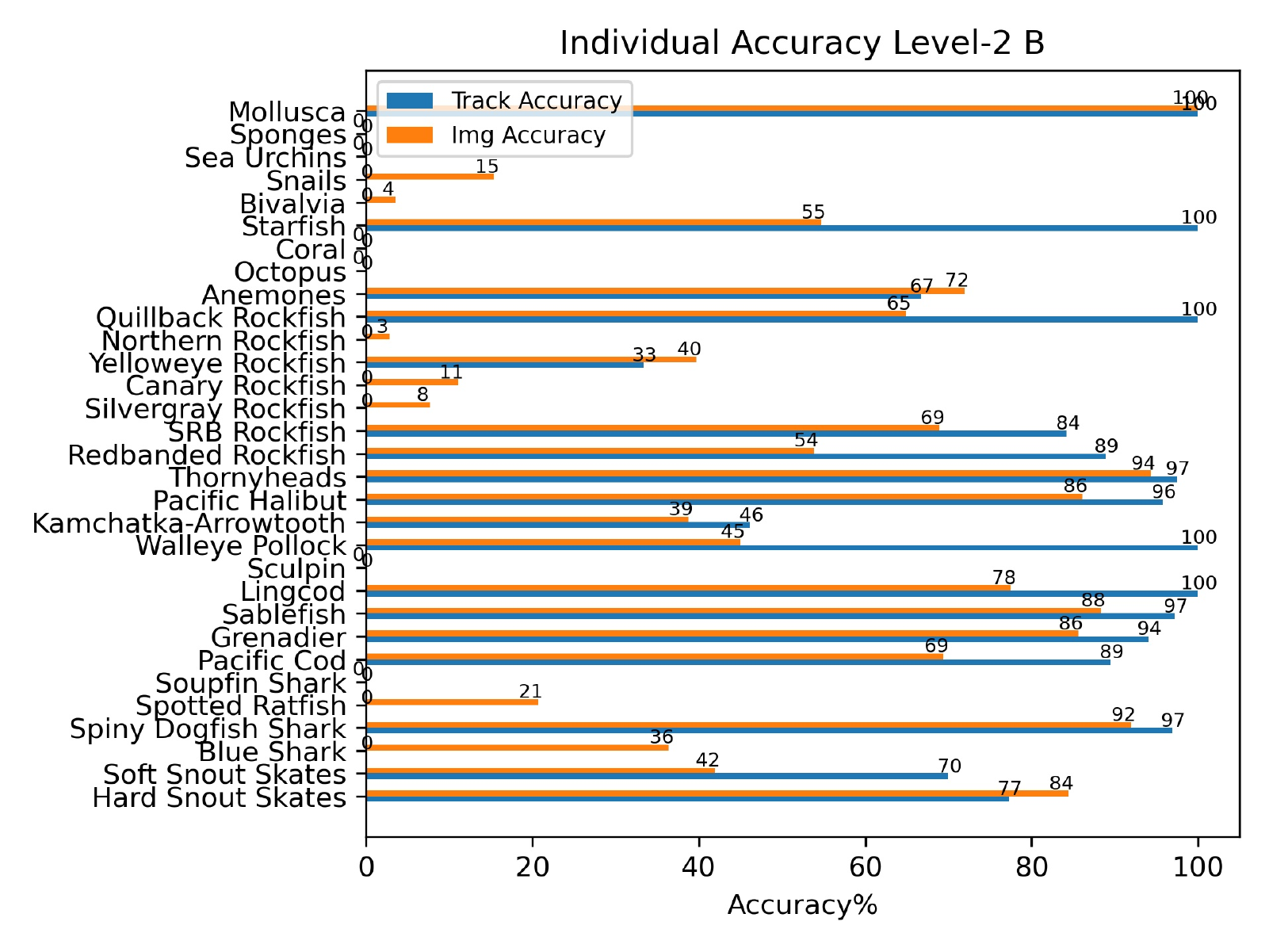}}
\subfigure[Percentage of Stopping at Level-1]{
\label{stop at level-1}
\includegraphics[width=0.48\linewidth]{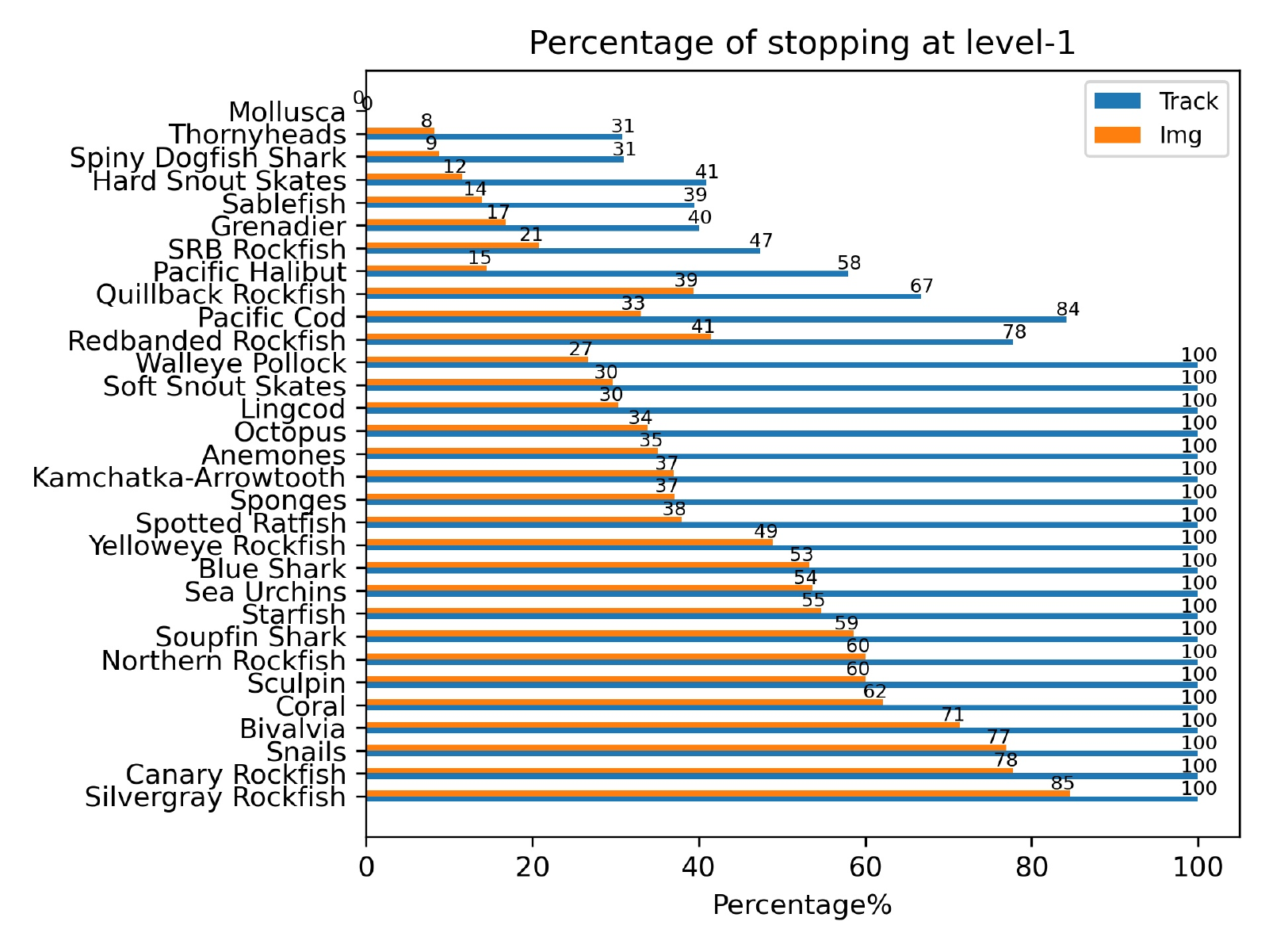}}
\caption{Detailed Accuracy on Coarse Level and Fine Level of the complete proposed system: The orange bar is image-based inference and the blue bar is video-based inference. (d) is under 'Level-2 C' evaluation method. We can see most tail species stop at coarse-level prediction, which makes $5.5\%$ improvement in overall video-based accuracy showed in Table~\ref{experiments}.} 
\label{detailed model-3}
\end{figure}

Detailed accuracy on the coarse level and fine level of Scheme-3 (our proposed complete system), based on the maximum average confidence score (denoted as '$video$' in Table~\ref{experiments}) is in Fig. \ref{detailed model-3}.

% \begin{figure}
%     \centering
%     \begin{minipage}[c]{0.49\textwidth}
%         \centering
%         \includegraphics[width= \linewidth]{Individual Accuracy-Model7 track based Level-1.pdf}
%         % \subcaption{}
%         \label{stereo}
%     \end{minipage}
%     \begin{minipage}[c]{0.49\textwidth}
%         \centering
%         \includegraphics[width= \linewidth]{Individual Accuracy-Model7 track based Level-2 A.pdf}
%         % \subcaption{}
%         \label{ours}
%     \end{minipage}
%     \begin{minipage}[c]{0.49\textwidth}
%         \centering
%         \includegraphics[width= \linewidth]{Individual Accuracy-Model7 track based Level-2 B.pdf}
%         % \subcaption{}
%         \label{brute force}
%     \end{minipage}
%     \begin{minipage}[c]{0.49\textwidth}
%         \centering
%         \includegraphics[width= \linewidth]{Model-7 percentage stop at level-1 track based.pdf}
%         % \subcaption{}
%         \label{subset}
%     \end{minipage}
%     \caption{Detailed Accuracy on Coarse Level and Fine Level of Model-3.}
%     \label{detailed model-3}
% \end{figure}

\subsection{Ablation Study}
Scheme-1 and Scheme-2 are experimented mainly for ablation study purposes. \textit{Comparing Scheme-1 with Scheme-2 from Table~\ref{experiments}, we can see confidence-score multiplication operations can effectively enforce hierarchical data structure and improve the performance even when Scheme-2 only trains two heads each time. Comparing Scheme-2 with Scheme-3, we can see our efficient training strategy ($Loss_2$) improves the performance by fully enforcing hierarchical data structure during training.}

% \begin{figure}
% \centering
% \includegraphics[width=0.7\textwidth]{Model-7 percentage stop at level-1 track based.pdf}
% \caption{This is from Model-3 under 'Level-2 D' evaluation method. The orange bar is image-based inference and the blue bar is video-based inference. Most tail species stop at coarse-level prediction. From Table~\ref{experiments}, it makes $5.5\%$ improvement in overall video-based accuracy.} 
% \label{stop at level-1}
% \end{figure}

Under 'Level-2 C', the completing systems' final predictions can stop at a coarse level if the final confidence score is lower than the greedy-searched threshold mentioned in the previous section. We call 'Level-2 C' as hierarchical prediction, which is one big advantage of the hierarchical classifier over flat classifiers, which allows fisheries managers to assign corresponding experts to review those images in a certain group and get the correct fine-level labels. \textit{Besides, from Fig.~\ref{stop at level-1}, we can see most tail-class species identification stop at a coarse level, resulting in a significantly higher overall accuracy in 'Level-2 C' over that of 'Level-2 B' in Table~\ref{experiments}. Also, our full system, Scheme-3, has the greatest number of images or tracks proceeding to fine level and  at the same time achieves the best performance.}

\section{Conclusions and Future Work}
We proposed an efficient hierarchical CNN classifier to enforce hierarchical data structure for fish species identification, combined with an efficient training strategy, and two video-based inference schemes. Our experiments show that the integrated use of these three main strategies indeed improves accuracy clearly. Additionally, hierarchical predictions allow images that cannot be confidently classified at the fine level to be confidently classified at a coarse level for experts future examination, which especially improve overall accuracy on tail-class species identification significantly by stopping at coarse-level predictions. Moreover, our method greatly outperforms the baseline method, a flat classifier. Future work will be devoted to adding more techniques like data sampling or additional training losses for tail-class species identification. It would be interesting to combine more strategies for long-tailed data with hierarchical classification.

\bibliographystyle{splncs04}
\bibliography{mybibliography}

\end{document}